\documentclass[twoside]{article}

\usepackage[accepted]{aistats2022}
\usepackage{apacite}
\usepackage{amsmath,amssymb,amsfonts}
\usepackage{algorithmic}
\usepackage{textcomp}
\usepackage{xcolor}
\usepackage{graphicx}
\usepackage{makecell}
\usepackage{changes}
\usepackage[breaklinks]{hyperref}
\usepackage{booktabs}
\usepackage{multirow}
\newcommand {\otoprule}{\midrule [\heavyrulewidth]} 
\newcolumntype {+}{ >{}}
 
\definechangesauthor[name={Wissam}, color=orange]{wis}
\definechangesauthor[name={Kanishka}, color=green]{Kan}
\definechangesauthor[name={Bach}, color=red]{Bach}
\usepackage[round]{natbib}

\begin{document}

%

%
\runningauthor{Van Bach Nguyen, Kanishka Ghosh Dastidar, Michael Granitzer, Wissam Siblini}
\twocolumn[

\aistatstitle{The Importance of Future Information in Credit Card Fraud Detection}

\aistatsauthor{ Van Bach Nguyen\\University of Duisburg-Essen \And Kanishka Ghosh Dastidar\\University of Passau

\AND Michael Granitzer\\University of Passau \And Wissam Siblini\\Worldline  }

\aistatsaddress{} 

]

\begin{abstract}
  Fraud detection systems (FDS) mainly perform two tasks: (i) real-time detection while the payment is being processed and (ii) posterior detection to block the card retrospectively and avoid further frauds. Since human verification is often necessary and the payment processing time is limited, the second task manages the largest volume of transactions. In the literature, fraud detection challenges and algorithms performance are widely studied but the very formulation of the problem is never disrupted: it aims at predicting if a transaction is fraudulent based on its characteristics and the past transactions of the cardholder. Yet, in posterior detection, verification often takes days, so new payments on the card become available before a decision is taken. This is our motivation to propose a new paradigm: posterior fraud detection with “future” information. We start by providing evidence of the on-time availability of subsequent transactions, usable as extra context to improve detection. We then design a Bidirectional LSTM to make use of these transactions. On a real-world dataset with over 30 million transactions, it achieves higher performance than a regular LSTM, which is the state-of-the-art classifier for fraud detection that only uses the past context. We also introduce new metrics to show that the proposal catches more frauds, more compromised cards, and based on their earliest frauds. We believe that future works on this new paradigm will have a significant impact on the detection of compromised cards.
\end{abstract}

\section{\uppercase{Introduction}}\label{sec:intro}

Credit card fraud is characterized by an entity getting unauthorized access to a card in order to make a purchase \citep{kou,Cheng,Lucas}. Although payment service providers are adopting prevention and detection tools to combat fraud, the evolving sophistication of fraudulent strategies continues to result in huge amounts of monetary losses \citep{Pozzolo3}. In 2018 alone, the total value of fraudulent transactions on cards issued in the SEPA area amounted to 1.8 billion Euros\footnote{\url{https://www.ecb.europa.eu/pub/cardfraud/html/ecb.cardfraudreport202008~521edb602b.en.html}, Last access: 12.10.2021}. 

Anti-fraud pipelines consist of a variety of components including fraud prevention, detection, and containment modules. Prevention methods \citep{adams} involve taking preemptive steps to reduce the likelihood of fraud occurring on a card through security tools and analysis of fraud trends.  Detection mechanisms monitor the incoming stream of transactions and operate in real-time or near real-time to either directly reject a transaction or raise alerts for human experts to further investigate \citep{Pozzolo2}. Additionally, fraud containment strategies aim to limit losses on breached cards by blocking them as soon as possible. Fraud detection mechanisms lie on a broad spectrum of approaches \citep{kou, Bhattacharyya,leborgne2022fraud}. On one end, we have rule-based approaches, devised and updated by domain experts \citep{giannini}. This requires human time, effort and maintenance, and cannot model very complex patterns. The other end is machine learning approaches that reduce this dependence on domain knowledge by learning patterns from the raw historical data in order to predict whether an incoming transaction is fraudulent or not \citep{awoyemi}. 

 In practice, discriminating accurately between the two classes based solely on raw information about the transaction is extremely difficult. What strongly characterizes a transaction as fraudulent is its discrepancy with respect to the consumer's habits, hence the need for contextualization with aggregation or representation of the cardholder's history of purchases \citep{Bahnsen, Ghosh, Jurgovsky}. To the best of our knowledge, available research studies on credit card fraud detection implement similar experimental protocols with a temporal split between training data and validation data, in which each sample is a transaction, sometimes contextualized with the past transactions of the cardholder. The novelty is usually proposed on the design of the algorithms, on a specific pre-processing or training method (sub-sampling, transfer learning) to tackle the known challenges of fraud detection (class imbalance, concept drift, class overlap, etc.). The regular experimental setting is natural for real-time fraud detection because the moment a transaction enters the system, only its information and its past are available. In this setting, algorithms like the LSTM \citep{Hochreiter2,Graves,Jurgovsky} have proven to be beneficial in practice by raising accurate alerts and making the best out of the verification bandwidth of human expert investigators.

However, because this bandwidth is limited, not all frauds are instantly detected. Data show that less than 1\% of fraudulent cards are confirmed by an expert within an hour and less than 30\% within a day (Figure~\ref{fig:verif_delay}). For the remaining 70\%, over the course of the day, 90\% of the cardholders make an additional transaction and 70\% make three new transactions, which could be used as well in a post-analysis for fraud detection. This is our motivation to propose, in this paper, a novel paradigm for posterior detection, as a complement to a standard real-time detection system. The question is the following: would a posterior model (exploiting ``future" transactions) improve the accuracy of alerts and catch extra frauds? Simple intuitions favor the ``yes" hypothesis. For instance, in the case of travel, the first transaction in the foreign country might raise an alert on a classical system but having the following transactions exclusively in the same area would discard the fraud suspicion. On the contrary, in the frequent case where frauds come in series, information about the next transactions might comfort the fraud likelihood.
 
To the best of our knowledge, this is the first time that the use of ``subsequent" transactions for posterior credit card fraud detection is studied. On a real-world dataset provided by a world leader in the payment industry, we start by making a statistical analysis of the time delay between transactions in each account to determine the appropriate quantity of ``future" information to use to have a real impact on compromised card detection. We then employ a Bi-LSTM to classify a transaction based on its characteristics and both the previous and next transactions from the cardholder. We compare it to previous state-of-the-art models such as Random Forest or LSTM, which only use the previous transactions. We also conduct several experiments to find out the most effective design for post-detection. In a nutshell, we make the following contributions:
\begin{itemize}
    \item We demonstrate that the time delta between transactions in each account is short enough, compared to the verification latency, so that the use of ``future" information has a practical application.
    \item We show that a Bidirectional LSTM can outperform the previous models like LSTM and Random Forest, showing the interest of posterior detection.
    \item We empirically observe that ``future" information is the most beneficial in the case where the past and the ``future" information are balanced in terms of the number of transactions considered.
    \item We enrich our evaluation of the approaches using two additional variants of the AUCPR metric: at the card and the ``early" card level. The latter, in particular, is a novel variant to evaluate the ability of a model to detect the first fraudulent transaction on a card.

\end{itemize}
With this work, we hope to motivate future research in the same direction. The paper is organized as follows: in the next section, we formalize the problem after analyzing the distribution of time delays between a cardholder's transactions. We then review classical state-of-the-art approaches for sequential credit-card fraud detection, as well as the applications of the Bi-LSTM in the general time-series tasks, before introducing in detail the design of our Bi-LSTM for credit card fraud detection. Next, we present our experimental setup and discuss its outcomes before concluding.

\section{\uppercase{Posterior Fraud Detection}}
A typical fraud detection system is divided into several modules that serve different purposes. The two main ones are \textbf{real-time detection} and \textbf{posterior detection}. The \textbf{first} is executed while the payment is being processed in order to block the transaction when it is suspicious. Because of processing time constraints, this module needs to be executed within a hundred milliseconds. Therefore, it usually relies on simple expert rules and machine learning models which make their decision based only on raw information about the transaction. Moreover, to avoid the inconvenience of blocking a genuine payment, this kind of system favors precision over recall. For these reasons, a large part of frauds remains to be caught afterward. This is where the \textbf{second} module comes into play. Using more advanced techniques with contextual features (computed from past transactions), posterior detection raises more accurate alerts after the transaction is completed, and these are verified by human investigators who block the card if the fraud is confirmed. Because the availability of investigators is limited, there is usually a verification delay that can range from few minutes to several days, depending on the risk level associated with the alert. We argue that during this delay, additional subsequent transactions are sometimes made with the same card, and these can also be used as context to improve detection. In the rest of the paper, we will use the terminology \textbf{``future transaction"} to refer to these transactions. 

With this consideration, we introduce a novel paradigm for fraud detection called posterior detection with future information. Its goal is to classify transactions with contextual features computed from both the preceding and following transactions from the same card. To properly determine if there is a real interest in such a paradigm, we measure, in this section, (1) the distribution of delays between the first fraud occurring on a card and the moment that the card is blocked and (2) the proportion of extra information available during this range of delays. 

Note that posterior detection with future information is not only interesting for unverified transactions that raised suspicion, but also to post analyze any other transaction that could have passed through the radar of a first detection system. Indeed, some unsuspected transactions are reported as fraudulent by cardholders themselves, sometimes weeks after they have been processed. Also, note that the objective of this novel paradigm is not to replace regular fraud detection but to complement it.

\subsection{Verification delay distribution}

\begin{figure}[h]
\centering
\includegraphics[scale = 0.12]{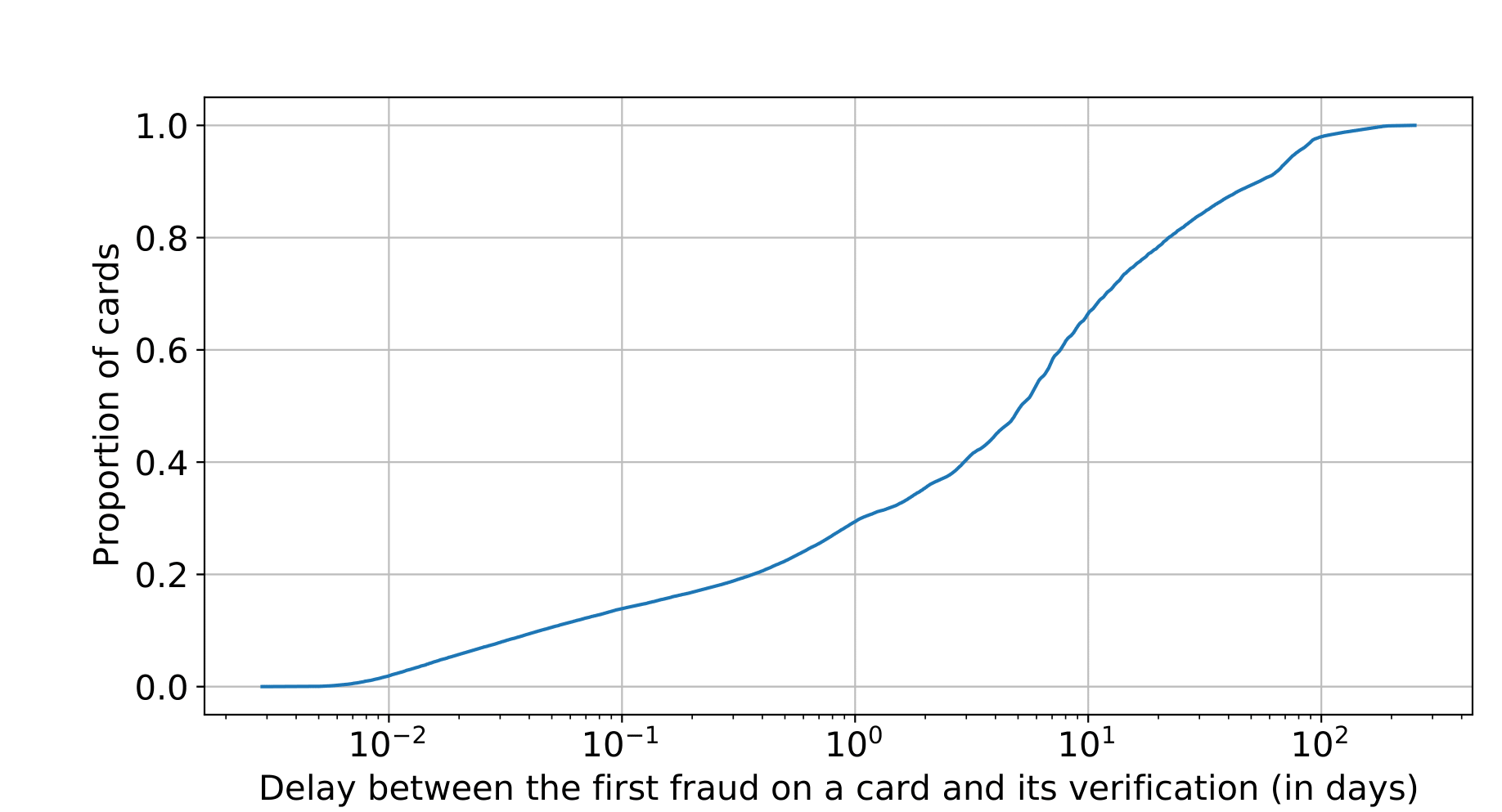}
\caption{Cumulative frequencies of compromised cards' verification delay.}
\label{fig:verif_delay}
\end{figure}

We here report verification delays measured on a real-world fraud detection system. As displayed in Figure~\ref{fig:verif_delay}, only 1\% of fraudulent cards are blocked within one hour after the first fraud occurs on their account and only 30\% within one day. A large majority are blocked within one week to ten days, which is why this delay is considered as the verification delay in most studies from the literature \citep{Pozzolo3,lebichot,alazizi}. 

To complement these statistics, we need to measure the average quantity of future information that becomes available faster than the verification latency.

\subsection{Availability of future transactions}

\begin{figure}[!htbp]
\centering
\includegraphics[scale=0.5]{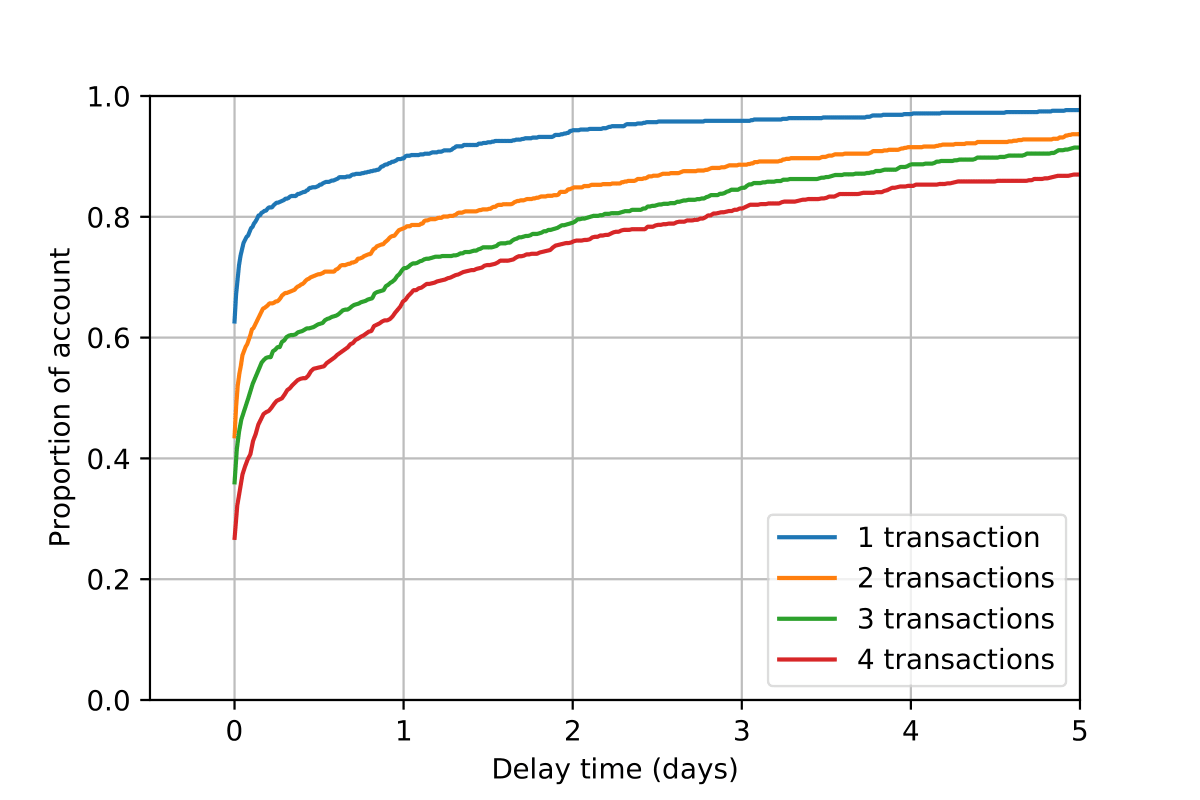}
\caption{Evolution of the proportion of cards that have made 1,2,3 or 4 transactions since the occurrence of their first fraud with respect to the corresponding delay.}
\label{fig:time_delay}
\end{figure}

Figure~\ref{fig:time_delay} shows the difference in time between the occurrence of the first fraudulent transaction and each of the four subsequent transactions. In particular, this cumulative graph illustrates that more than 90\% of accounts have at least one extra transaction within one day. And for approximately 78\%, 70\%, and 66\% of the accounts, the next two, three, and four transactions are respectively executed within the same day.
The beginning of the curve is interesting to analyze: all the curves directly jump from zero to a significant value. This means that a large percentage of frauds are followed by one, two, three, or even four transactions after a very short delay. This ties in with an expert intuition that fraudsters repetitively exploit cards that they gain access to within a limited time budget, to maximize their gains. 

By crossing information from Figure~\ref{fig:time_delay} and Figure~\ref{fig:verif_delay}, we can see that there is a non-negligible proportion of compromised cards that are still undetected after one day, and on which at least three extra transactions are available. In fact, this number lies between $40\%$ and $70\%$ of the breached cards. This means that a more accurate post-detection system speeding up their detection would entail non-negligible financial savings\footnote{Increasing the ``risk" score of a still unverified compromised card changes the verification priority and allows catching it faster.}.

\subsection{Problem formulation}\label{sub:formulation}

Let us here formally define the problem of fraud detection with future information. Consider a set of transactions $S_t$, where each element $t_i$ is associated to a card $c_i$ and a label $l_i$. For a transaction $t_i \in S_t$, let us denote by $P_i$ (resp. $F_i$) the set of $m_P$ (resp. $m_F$) transactions in $S_t$, associated to the same $c_i$, and chronologically preceding (resp. following) $t_i$. The objective of a fraud detection with future information is to predict the label $l_i$ of $t_i$ from the features of $t_i$, $P_i$ and $F_i$. And the important question is to determine whether a system that uses $t_i$, $P_i$, and $F_i$ can outperform a system that only uses $t_i$ and $P_i$? In a more intuitive way, could future context improve fraud detection? Obviously, several questions are naturally associated with the main one: what is the optimal future context size $m_F$ so that the system has a practical use (both good performance and short enough delay)? What architecture design should we choose for our model to be able to use both past and future contexts? In what proportion ($m_P$ and $m_F$) should each context be exploited?

\section{\uppercase{Related Work}}
The credit-card fraud detection problem studied in the literature exclusively consists in predicting the fraud label $l_i$ from either $t_i$ alone (context-free approaches) or $t_i$ and $P_i$ (context-based approaches). It primarily focuses on discriminant approaches  \citep{Bhattacharyya, Cheng, Lucas} that aim at improving the accuracy of detection. This often breaks down to tackling specific peculiarities of the problem such as concept drift \citep{Pozzolo}, class imbalance \citep{Pozzolo2} and the selection of appropriate metrics \citep{Siblini}. 

Context-free approaches generally underperform and only identify simple fraud patterns (e.g. transactions carried out on compromised terminals in risky areas) \citep{Jurgovsky}, because of the significant overlap between the classes when using raw features only. Recent popular approaches are almost exclusively context-based. A myriad of different ways have been proposed in the literature to tackle the problem of representing a user's past context $P_i$. On the one hand, a popular strategy consists of using manual feature aggregates. This was first proposed in \cite{Whitrow}, where features relevant to fraud detection are identified and transactions are grouped based on these features. Different statistics such as the sum or average of amounts are then calculated for these groups. \cite{Bahnsen} expanded this through aggregates that use multiple features as conditions for grouping transactions. On the other hand, several works have proposed to automatically represent this context from the history of transactions without the need for manual intervention. These include the use of convolutional neural networks \citep{Fu} or the recent work by \cite{Ghosh} that generates aggregates for e-commerce transactions with an end-to-end neural network-based method that learns a set of condition-features and aggregation functions to optimize fraud detection. For card-present data, the state-of-the-art is the LSTM \citep{Jurgovsky}. The latter is usually trained on sequences of either 5 or 10 transactions grouped chronologically according to user accounts. Comparisons with classical methods like Random Forest where the user's context is represented through manual feature aggregations, as in \cite{Bahnsen}, show that both approaches perform comparably. The advantage of the LSTM is that it lowers the need for business expertise.

In our application scenario, i.e. posterior fraud detection, one of the limitations of manual aggregates and the LSTM is that they only take into account the past or preceding information in order to build the context for the current data point. Instead, we could use the Bi-directional LSTM (Bi-LSTM)  proposed by \cite{Graves} which uses both left/past and right/future context and was shown to outperform the LSTM, RNN, and Bi-RNN for speech recognition on the TIMIT speech database. It also has been used in several time-series tasks such as energy load forecasting \citep{Persio}, stock market prediction \citep{Althelaya} and network-wide traffic speed prediction \citep{Cui}. In the fraud detection domain, Bi-LSTM has already been applied as a baseline to detect fraudulent transactions \citep{li2019}. However, \cite{li2019} only extracts the last state data as a user embedding without considering the succeeding transactions. Thus far, Bi-LSTM that makes use of future transactions  $F_i$ has not been proposed for the credit card fraud detection problem.  





\section{BI-LSTM FOR POSTERIOR FRAUD DETECTION}

In our problem, each input ($P_i$,$t_i$,$F_i$) is a sequence of transactions with the past ones, the current one, and the future ones, ordered chronologically. As credit card fraud data conforms to a tabular format, each transaction $t_i$ can be considered as a vector of features, that can be either numerical or categorical.

\begin{figure}[h]
\centering
\includegraphics[scale= 0.65]{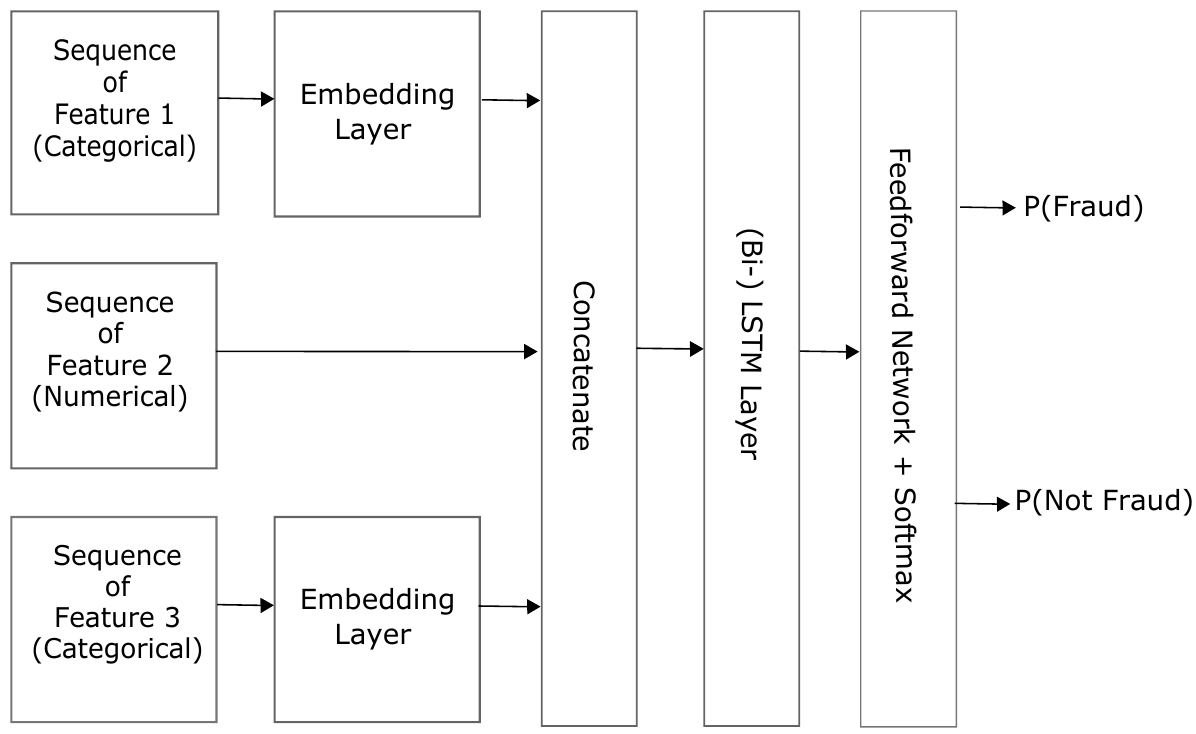}
\caption{The general pipeline of LSTM and Bi-LSTM for fraud detection.}
\label{fig:LSTM_architecture}
\end{figure}

We describe here the global design of our approach, based on the Bi-LSTM, which takes as input the aforementioned sequence. Figure~\ref{fig:LSTM_architecture} illustrates the overall architecture. The latter also applies to a regular LSTM that takes as input the sequence with only past and current information ($P_i$, $t_i$). The rest of the section describes in detail each step in the pipeline.

\subsection{Preprocessing and Feature Engineering}
Transactions are described by 30 features. Numerical features, such as the ``amount", are clipped into the range [0,1000] to remove outliers, then standardized to a mean $\mu = 0$ and a standard deviation $\sigma = 1.0$. For the deep learning models like the LSTM and Bi-LSTM, categorical features are transformed into continuous vectors. For this, we use Word2Vec applied to the sequences of discrete values taken by the feature for all card-holders as proposed in \cite{Russac}. Each value of a categorical feature $f$ is finally translated into a vector $\mathbf{y}^{f} \in \mathbb{R}^{d_{f}}$ \citep{Ghosh}.

Besides the raw features, we additionally use aggregated features \citep{Bahnsen,Jurgovsky,Ghosh}. This serves as a feature engineering method to capture the past context of a card-holder. In short, information from previous transactions which respect certain feature constraints (e.g. take place in the same country as the current transaction), is \textit{aggregated} with an operator like \textit{sum} or \textit{count}.
An aggregated feature can be the sum of amounts of all transactions that took place in the same country within the last 24 hours.
For a more in-depth insight, we refer the readers to \cite{Bahnsen}. In our experiments, we use three functions \textit{sum, mean, count} to aggregate the information of past transactions within 24 hours, using several constraint features such as the country or the merchant category code. We obtain 13 aggregated features in total. In the following, we will refer to a feature set with only raw features as the $Base$ feature set and the combination of raw and aggregated features as the $Base+Agg$ feature set.

Since LSTM and Bi-LSTM take a sequence as input, sequence extraction from the raw set of transactions is necessary.  As in \citep{Jurgovsky}, we construct the list of sequences with a fixed length $l_s$  (e.g 5) from each user by ordering chronologically all of its transactions and by sliding a window with a stride of one. We obtain $n_s = n_c-l_s+1$ sequences for a user $c$, in which, $n_c$ is the total number of transactions of the user. For example, if a user has seven transactions $1,2, \ldots , 7$ and sequence length $l_s$ is 5, then the sequence list of this user is $[1,2,3,4,5],[2,3,4,5,6],[3,4,5,6,7]$. We finally merge the sequence lists of all users to constitute the final dataset for the model.
\subsection{Bidirectional LSTM architecture}
\begin{figure}[h]
\centering
\includegraphics[scale=0.5]{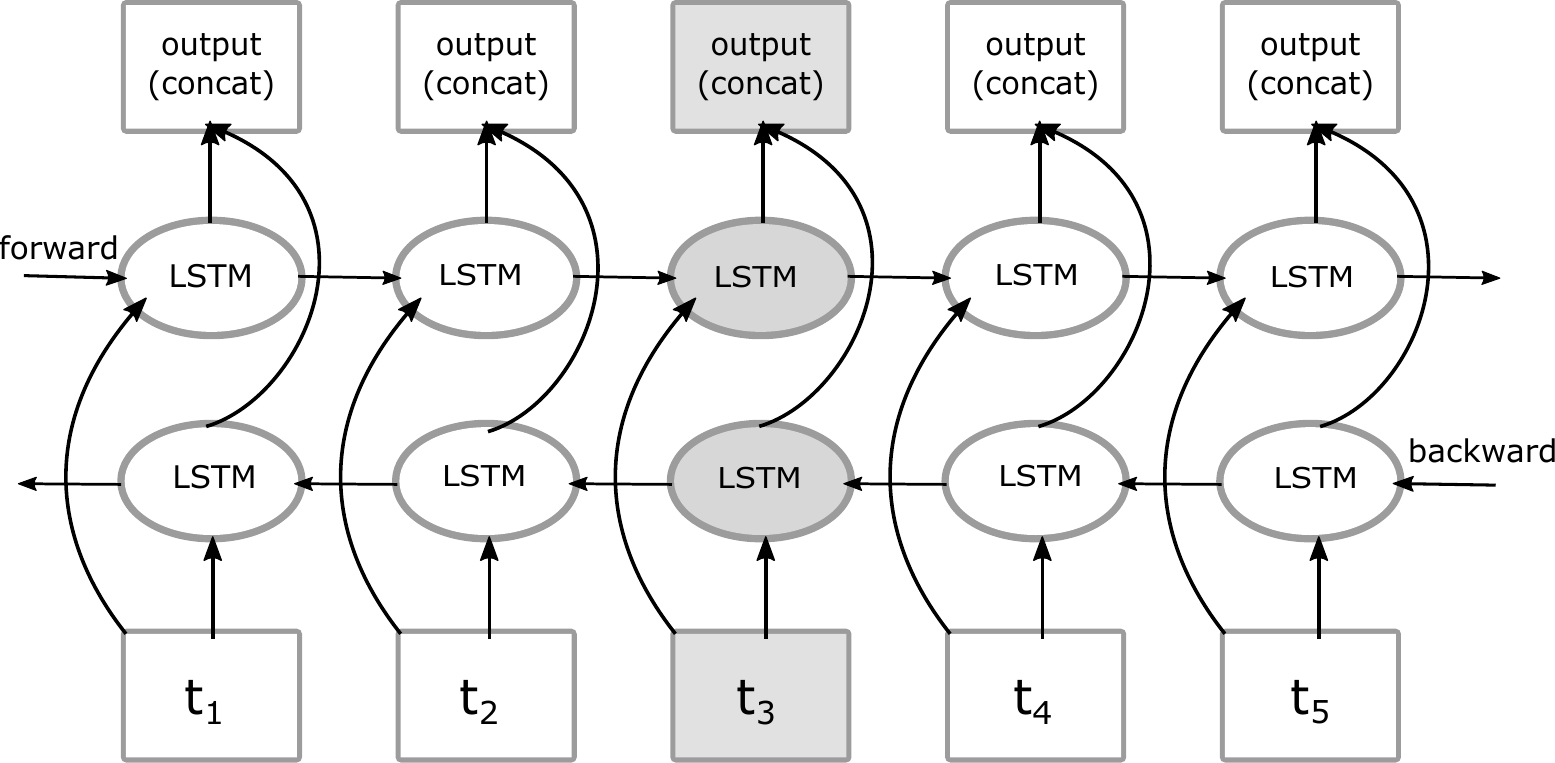}
\caption{Bi-LSTM architecture: $t_i$ is the transaction at step $i$ in the sequence. Gray color indicates the target transaction.}
\label{fig:BiLSTM}
\end{figure}
\begin{table*}[ht]
\centering
\caption{Comparison, in terms of the three AUCPR scores, of a Random Forest and variants of LSTM/Bi-LSTM trained with the two sets of features. Compared to the baseline LSTM 4-1-0 with sequence length 5, LSTM 6-1-0 and Bi-LSTM 4-1-2 respectively add two transactions from the past and two from the future.}
\begin{tabular}{@{}cccccc@{}}
\toprule
\multicolumn{1}{l}{}                                                                              &                                 & \begin{tabular}[c]{@{}c@{}}Random Forest \\ 0-1-0\end{tabular} & \begin{tabular}[c]{@{}c@{}}LSTM \\  4-1-0\end{tabular} & \begin{tabular}[c]{@{}c@{}}LSTM \\ 6-1-0\end{tabular} & \begin{tabular}[c]{@{}c@{}}\textbf{Bi-LSTM}\\ \textbf{4-1-2}\end{tabular} \\ \midrule
\multicolumn{1}{c|}{\multirow{2}{*}{\begin{tabular}[c]{@{}c@{}}Transaction\\ Level\end{tabular}}} & \multicolumn{1}{c|}{Base}       & \multicolumn{1}{c|}{0.163 ± 0.006}                             & \multicolumn{1}{c|}{0.313 ± 0.011}                     & \multicolumn{1}{c|}{0.330 ± 0.014}                    & \textbf{0.367 ± 0.008}                                  \\
\multicolumn{1}{c|}{}                                                                             & \multicolumn{1}{c|}{Base + Agg} & \multicolumn{1}{c|}{0.219 ± 0.003}                             & \multicolumn{1}{c|}{0.344 ± 0.007}                     & \multicolumn{1}{c|}{0.356 ± 0.017}                    & \textbf{0.403 ± 0.021}                                  \\ \midrule
\multicolumn{1}{c|}{\multirow{2}{*}{\begin{tabular}[c]{@{}c@{}}Card Level \\ Early\end{tabular}}}       & \multicolumn{1}{c|}{Base}       & \multicolumn{1}{c|}{0.061 ± 0.007}                             & \multicolumn{1}{l|}{0.141 ± 0.009}                     & \multicolumn{1}{l|}{0.157 ± 0.014}                    & \multicolumn{1}{l}{\textbf{0.172 ± 0.007}}              \\
\multicolumn{1}{c|}{}                                                                             & \multicolumn{1}{c|}{Base + Agg} & \multicolumn{1}{c|}{0.110 ± 0.003}                             & \multicolumn{1}{l|}{0.165 ± 0.009}                     & \multicolumn{1}{l|}{0.172 ± 0.017}                    & \multicolumn{1}{l}{\textbf{0.254 ± 0.012}}              \\ 
\midrule
\multicolumn{1}{c|}{\multirow{2}{*}{\begin{tabular}[c]{@{}c@{}}Card Level \\ Regular\end{tabular}}}       & \multicolumn{1}{c|}{Base}       & \multicolumn{1}{c|}{0.201 ± 0.005}                             & \multicolumn{1}{l|}{0.342 ± 0.019}                     & \multicolumn{1}{l|}{0.358 ± 0.012}                    & \multicolumn{1}{l}{\textbf{0.403 ± 0.008}}              \\
\multicolumn{1}{c|}{}                                                                             & \multicolumn{1}{c|}{Base + Agg} & \multicolumn{1}{c|}{0.231 ± 0.005}                             & \multicolumn{1}{l|}{0.359 ± 0.012}                     & \multicolumn{1}{l|}{0.372 ± 0.011}                    & \multicolumn{1}{l}{\textbf{0.417 ± 0.017}}              \\ 

\bottomrule
\end{tabular}
\label{tab:AUCPR_F2F_add}
\end{table*}
The Bi-directional Long Short-Term Memory (Bi-LSTM) concatenates the outputs of the two LSTMs \citep{Hochreiter2} and feeds the result to a fully connected layer to finally predict the class (fraud or genuine) scores. Figure~\ref{fig:BiLSTM} illustrates the Bi-LSTM architecture. In the figure, $t_i$ denotes a transaction at position $i$. The gray transaction is the target transaction for which we want to predict the label. In our experiments, we will compare the Bi-LSTM with the LSTM (and other approaches) based on their predictions on the same sets of transactions. We will use, as much as we can, the same parameters and data preparation (description above) for both models. $m_F$ and $m_P$ are hyperparameters that we will adjust, by changing the target transaction, based on the purpose of the experiments. This means the target transaction will not necessarily be $t_3$ as in Figure~\ref{fig:BiLSTM}.


\section{\uppercase{Experimental Setup}}
\label{sec:expe}
Our experiments aim at answering the problem formulated in section \ref{sub:formulation}. We here describe our design, i.e. the dataset and features, the compared classifiers, and the metrics used. For a fair comparison with the state-of-the-art LSTM, our experimental setup will inherit from \cite{Jurgovsky}.
\paragraph{Dataset}
Credit card data are highly confidential, so there are almost no high-quality public datasets to consider in this study. The most commonly used one is a Kaggle dataset\footnote{\url{https://www.kaggle.com/mlg-ulb/creditcardfraud}, Last access: 12.10.2021.}. It only consists of around 300,000 anonymized transactions and the features are PCA-transformed. Since the transactions are not associated with users and the features are transformed, the dataset is unsuitable for our experiments. Another public dataset is PaySim\footnote{\url{https://www.kaggle.com/ealaxi/paysim1}, last access: 12.10.2021.}, which is synthetic and related to mobile fraud detection. However, the characteristics of mobile payment transactions are very different from the card-present transactions. Therefore, we consider instead a real-world fraud detection system from our industrial partner, a world leader in the payment industry. Our private dataset contains transactions from clients of various banks from January 2017 to April 2017. Each transaction is labeled by a team of human experts. The two classes are highly imbalanced as only 0.0263\% of the 30 million transactions are fraudulent. To deal with this imbalance, we apply the \textit{account sampling} technique from the previous work \citep{Jurgovsky}. 
\paragraph{Classifiers}
We select the classifiers for the experiments according to the questions we aim at answering. The first classifier is Random Forest, which is a context-free approach that uses only the present transaction. Random Forest is a common baseline that has been used in several studies \citep{Jurgovsky,Pozzolo,Ghosh,Lucas}. We can consider the Random Forest to be a context-based approach when applied to the $Base+Agg$ feature set. The second classifier that we use is the LSTM - the state-of-the-art sequential classifier- which utilizes the past context to predict the current transaction. Finally, we consider our proposal, the Bi-LSTM, which takes into account the past, the present, and future information. For training both the LSTM and Bi-LSTM, we use the cross-entropy loss function, and the Adam optimizer \citep{kingma2014adam}. 
\paragraph{Validation}
To select the best hyperparameters for each classifier, we hold the most recent part of the training set as validation and use a random search over 40 iterations on a pre-defined grid of hyperparameters values to maximize the AUCPR score on it. For the random forest, we look for the optimal number of trees in the forest and tree depth. For the LSTM and the Bidirectional LSTM we look for the optimal hidden state size, layers size in the feedforward network, drop out rate, learning rate, number of epochs, and batch size. The detail of hyperparameters for each model can be found in Appendix~\ref{sec:apdx:hyper}, Table~\ref{apdx:tab:hyper}.

\paragraph{Evaluation}

\begin{table*}[h]
\centering
\caption{Comparison between three variants of the Bi-LSTM (different number of future/past transactions) with the same sequence length of 5. The LSTM with the same sequence length is also added as a baseline.}
\begin{tabular}{@{}cccccc@{}}
\toprule
\multicolumn{1}{l}{}                                                                              &                                 & \begin{tabular}[c]{@{}c@{}}LSTM \\  4-1-0\end{tabular} & \begin{tabular}[c]{@{}c@{}}Bi-LSTM \\  3-1-1\end{tabular} & \begin{tabular}[c]{@{}c@{}}\textbf{Bi-LSTM} \\  \textbf{2-1-2}\end{tabular} & \begin{tabular}[c]{@{}c@{}}Bi-LSTM\\ 1-1-3\end{tabular} \\ \midrule
\multicolumn{1}{c|}{\multirow{2}{*}{\begin{tabular}[c]{@{}c@{}}Transaction\\ Level\end{tabular}}} & \multicolumn{1}{c|}{Base}       & \multicolumn{1}{c|}{0.313 ± 0.011}                             & \multicolumn{1}{c|}{0.322 ± 0.006}                     & \multicolumn{1}{c|}{\textbf{0.351 ± 0.007}}                  & 0.323 ± 0.005                                  \\
\multicolumn{1}{c|}{}                                                                             & \multicolumn{1}{c|}{Base + Agg} & \multicolumn{1}{c|}{0.336 ± 0.006}                             & \multicolumn{1}{c|}{0.349 ± 0.017}                     & \multicolumn{1}{c|}{\textbf{0.381 ± 0.007}}                    & {0.370 ± 0.007}                                  \\ \midrule
\multicolumn{1}{c|}{\multirow{2}{*}{\begin{tabular}[c]{@{}c@{}}Card Level \\ Early\end{tabular}}}       & \multicolumn{1}{c|}{Base}       & \multicolumn{1}{c|}{0.126 ± 0.006}                             & \multicolumn{1}{l|}{0.132 ± 0.005}                     & \multicolumn{1}{l|}{\textbf{0.160 ± 0.013}}                    & \multicolumn{1}{l}{{0.139 ± 0.011}}              \\
\multicolumn{1}{c|}{}                                                                             & \multicolumn{1}{c|}{Base + Agg} & \multicolumn{1}{c|}{0.137 ± 0.015}                             & \multicolumn{1}{l|}{0.182 ± 0.007}                     & \multicolumn{1}{l|}{\textbf{0.218 ± 0.010}}                    & \multicolumn{1}{l}{0.205 ± 0.014}              \\ 
\midrule
\multicolumn{1}{c|}{\multirow{2}{*}{\begin{tabular}[c]{@{}c@{}}Card Level \\ Regular\end{tabular}}}       & \multicolumn{1}{c|}{Base}       & \multicolumn{1}{c|}{0.326 ± 0.019}                             & \multicolumn{1}{l|}{0.347 ± 0.007}                     & \multicolumn{1}{l|}{\textbf{0.377 ± 0.011}}                    & \multicolumn{1}{l}{0.317 ± 0.016}         \\
\multicolumn{1}{c|}{}                                                                             & \multicolumn{1}{c|}{Base + Agg} & \multicolumn{1}{c|}{0.340 ± 0.009}                             & \multicolumn{1}{l|}{0.360 ± 0.013}                     & \multicolumn{1}{l|}{\textbf{0.388 ± 0.011} }                   & \multicolumn{1}{l}{0.376 ± 0.011}              \\ 

\bottomrule
\end{tabular}
\label{tab:AUCPR_F2F_5}
\end{table*}
In our experiments, we use 2 months of transactions (24.01-24.03) for training, one week (25.03-31.03) of gap, and one month for testing (01.04-30.04). This data-split strategy is based on production requirements as it takes time for human experts to label the transactions \citep{Pozzolo3,lebichot,alazizi}, hence the one-week gap between the train and test set. For the metrics, we use the Area Under Precision-Recall curve (AUCPR), which has been shown to be more suitable than the AUC-ROC for highly imbalanced datasets \citep{SR15}. We report three different AUCPR-based metrics for each experiment: at the transaction level, at the card level, and early at the card level.

At the \textbf{transaction level}, we take into account all the transactions of the test data and calculate the AUCPR score on this set. This metric has been used in most of the recent works on credit card fraud detection \citep{Ghosh,lebichot,Lucas}. However, in practice, when an alert for a particular transaction is raised, investigators check the other transactions from the associated card. Therefore, it makes sense to compute the AUCPR on the set of cards instead of the set of transactions, which intuitively tells us if compromised cards tend to appear earlier than genuine cards with respect to transactions risk levels. More formally, to compute this metric, we consider the ground-truth label of a card as genuine if it contains no fraudulent transactions, and as fraud otherwise. As for the predicted score for the card, we consider the highest predicted score among its transactions. We then calculate the regular AUCPR for this set of card scores and card ground truth labels. 

We finally introduce an \textbf{AUCPR ``early''} at the \textbf{card-level}. The reason is that, in practice, it is beneficial to detect the first fraudulent transaction that occurred on a card in order to track the fraud history but also block the card as soon as possible. \textbf{AUCPR ``early''} is almost the same as the AUCPR at the card level except that, for compromised cards, instead of considering the transaction with the highest score, we consider the score predicted for the first fraudulent transaction on the card. The intuition here is that a high value for this metric means that the model tends towards scoring the first fraudulent transaction of a card higher than the maximal scores of genuine cards. And this leads to models that detect cards from their earliest fraud, and therefore to mitigation of losses. To avoid random errors, for each model and test set, we will report each result as the mean evaluation across 5 runs.
\section{\uppercase{Results and Discussion}}
\label{sec:result}
\begin{table*}[ht!]
\centering
\caption{Comparison between LSTM with a longer sequence (10) and two Bi-LSTM (2-1-2 and 4-1-2) with shorter sequences (5 and 7).}
\begin{tabular}{@{}ccccc@{}}
\toprule
\multicolumn{1}{l}{}                                                                              &                                 & \begin{tabular}[c]{@{}c@{}}LSTM \\  9-1-0\end{tabular}  & \begin{tabular}[c]{@{}c@{}}Bi-LSTM \\  2-1-2 

\end{tabular} & \begin{tabular}[c]{@{}c@{}}\textbf{Bi-LSTM} \\  \textbf{4-1-2} 

\end{tabular} \\ \midrule
\multicolumn{1}{c|}{\multirow{2}{*}{\begin{tabular}[c]{@{}c@{}}Transaction\\ Level\end{tabular}}} & \multicolumn{1}{c|}{Base}       & \multicolumn{1}{c|}{0.345 ± 0.015}                                                & \multicolumn{1}{c|}{0.357 ± 0.005} & \multicolumn{1}{c}{\textbf{0.374 ± 0.012}}                                                   \\
\multicolumn{1}{c|}{}                                                                             & \multicolumn{1}{c|}{Base + Agg} & \multicolumn{1}{c|}{0.368 ± 0.018}                                                  & \multicolumn{1}{c|}{0.368 ± 0.008} & \multicolumn{1}{c}{\textbf{0.433 ± 0.019}}                                                      \\ \midrule
\multicolumn{1}{c|}{\multirow{2}{*}{\begin{tabular}[c]{@{}c@{}}Card Level \\ Early\end{tabular}}}       & \multicolumn{1}{c|}{Base}       & \multicolumn{1}{c|}{0.203 ± 0.019}                                                  & \multicolumn{1}{l|}{0.218 ± 0.009} & \multicolumn{1}{c}{\textbf{0.231 ± 0.018}}                                 \\
\multicolumn{1}{c|}{}                                                                             & \multicolumn{1}{c|}{Base + Agg} & \multicolumn{1}{c|}{\textbf{0.289 ± 0.008}}                                                  & \multicolumn{1}{l|}{0.288 ± 0.012} & \multicolumn{1}{c}{0.272 ± 0.013}                                \\ 
\midrule
\multicolumn{1}{c|}{\multirow{2}{*}{\begin{tabular}[c]{@{}c@{}}Card Level \\ Regular\end{tabular}}}       & \multicolumn{1}{c|}{Base}       & \multicolumn{1}{c|}{0.384 ± 0.014}                                                  & \multicolumn{1}{l|}{0.414 ± 0.014} & \multicolumn{1}{c}{\textbf{0.439 ± 0.011}}                             \\
\multicolumn{1}{c|}{}                                                                             & \multicolumn{1}{c|}{Base + Agg} & \multicolumn{1}{c|}{0.415 ± 0.010}                                                  & \multicolumn{1}{l|}{0.411 ± 0.017 } & \multicolumn{1}{c}{\textbf{0.426 ± 0.016}}                                \\ 
\bottomrule
\end{tabular}
\label{tab:AUCPR_long_short}
\end{table*}
The following results and discussion are organized based on the questions mentioned in section \ref{sub:formulation}. To answer, we compare the classifiers under several forms. In particular, we associate each model to a triplet \textit{past-present-future} of digits denoting the number of past, present, and future transactions that they consider. For example, 3-1-1 indicates that the input of the model is a sequence of 5 transactions: 3 past transactions, 1 present transaction, and 1 future transaction. The target transaction in this particular case would be the 4th transaction (see Figure~\ref{fig:BiLSTM}). The number of present transactions is always 1 since we only predict one transaction at a time. For fair comparisons, models are compared on their predictions for the same set of ``present'' transactions. However, results across tables might not be comparable: since the approach with a maximal size for past context and future context differ from one table to the other, the set of transactions with enough subsequent and preceding transactions on which the models can be evaluated differ as well. For instance, in table 1, the maximum length is 7 (6-1-0), the training (testing) set contains 8.6M (3.7M) transactions with 0.7M (0.4M) accounts. Meanwhile, in table 2, the maximum length is 5, training (testing) sizes are 10.5M (5.0M) transactions with 1.0M (0.7M) accounts.

\textbf{Does adding future information improve predictive performance? Is this improvement more significant than the one entailed by adding extra information from the past?} To answer, we carry a first experiment where use a Random Forest and an LSTM 4-1-0 as baselines, and we add two classifiers. One with two additional transactions from the past, and one with two additional transactions from the future, i.e. an LSTM 6-1-0 and a Bi-LSTM 4-1-2. Results in table \ref{tab:AUCPR_F2F_add} first show, as we could expect, that adding more information from the past improves the performance of the model. However, adding two subsequent transactions (Bi-LSTM 4-1-2) is clearly more beneficial than adding two preceding transactions (LSTM 6-1-0). The higher performance across all metrics points us to three complementary conclusions. Firstly, the transaction level metric shows that the Bi-LSTM 4-1-2 is globally able to identify more fraudulent transactions. Secondly, the regular card-level metric shows that the Bi-LSTM is also able to identify additional fraudulent cards, which means that not all extra frauds caught are from the same cardholders and that there is a real added value for investigators. Lastly, the card-level early metrics show that the Bi-LSTM is able to catch more fraudulent cards at the earliest stage, i.e. after the first fraudulent transaction is made. This last advantage might not manifest in practice, since it can be counterbalanced by the fact that the Bi-LSTM 4-1-2 has to wait for the next two transactions anyway. Nevertheless, it is always interesting to track the first fraud in post-analysis to avoid mislabeling \citep{alazizi} since test data becomes training data in the future. Note that we also tried the Random Forest with aggregates on the 4-1-0 and 4-1-2 settings (by concatenating the sequence transactions into a single vector) and the AUCPR was better in the 4-1-2 setting (0.258±0.002) than in the 4-1-0 setting (0.230±0.002) and the 0-1-0 setting (0.219 ± 0.003), but it was still worse than the Bi-LSTM 4-1-2 (0.403 ± 0.021). This confirms that future information is useful for several models, but the Bi-LSTM is able to exploit it in a better manner than the Random Forest.

The use of aggregated features improves the performance in every case. This suggests the general usefulness of feature engineering and also that aggregates and sequential models pick up on different patterns from the past context and are complementary. The improvement when using aggregates is especially pronounced for detecting the first frauds. 

We observe that both sequence models are able to outperform Random Forest. Therefore, for the next experiments, we drop it and use only the LSTM as the baseline.

\textbf{In what proportion should the past and future contexts be exploited?}
In a second experiment, we keep the baseline LSTM with a sequence length of 5 with 4 past transactions (4-1-0). For the Bi-LSTM, we consider the same sequence length of 5 but we test different combinations sizes for past and future context. We consider two imbalanced scenarios, either with more future transactions (1-1-3) or with more past transactions (3-1-1). Additionally, we consider the scenario where the past and future contexts are balanced (2-1-2). 

Table~\ref{tab:AUCPR_F2F_5} shows that the balanced model achieves the highest performance across both feature sets and all three metrics. This suggests that is better to have a long enough context from both sides to compute relevant hidden states than to have a long future context or a long past context. We note a significant improvement of model 1-1-3 when using the aggregated features, almost to the level of model 2-1-2. This could be attributed to fact that the lack of past context in 1-1-3 is compensated by the aggregates.

\textbf{Can the Bi-LSTM use shorter sequences than the regular LSTM while remaining competitive in performance?}
In the first experiment, increasing the sequence length (LSTM 6-1-0 vs 4-1-0) seemed to have a positive impact on the LSTM's performance. 

Here we compare an LSTM with a large sequence length (9-1-0) with two Bi-LSTMs with a much shorter past context (2-1-2 and 4-1-2) and a shorter sequence overall. Results from table \ref{tab:AUCPR_long_short} show that we can have a sequence twice as short (2-1-2 vs 9-1-0) but the use of future information still leads to (Base features) or similar results (Base + Agg features). Increasing the past sequence length a little more with the Bi-LSTM (4-1-2) allows to further improve the performance, making it more competitive than the LSTM, despite the latter's longer sequence. Having a shorter sequence and past context is interesting. Since the model requires the number of available transactions for the accounts to be equal to the sequence length in order to be optimal, the longer the sequence is, the fewer accounts can be considered. 

\section{\uppercase{Conclusion}}

In this paper, we investigated a new research direction for credit card fraud detection. Whereas most of the papers focus on real-time or near real-time fraud detection to predict the label of a transaction, we chose, motivated by a verification delay that allows it, to explore the use of future transactions in order to improve detection. With a relevant Bi-LSTM design, we showed that using only two transactions from the future, usually available within a day, we can build a post-detection system that catches more fraudulent transactions, more fraudulent cards, and at an earlier stage. We note however that our proposal is not fit for real-time detection. They are made to complement classical Fraud Detection Systems in post-detection. With this paper, we hope to encourage further proposals in this direction with more and more accurate algorithms. Furthermore, our Bi-LSTM could be extended to integrate an attention mechanism. In our case, we plan to analyze the problem from the interpretability point of view. In particular, our idea is to extend recent works such as the Neural Aggregate Generator from \cite{Ghosh} to introduce the use of the future context. This would allow getting insights on the kind of aggregations from the future that is relevant for fraud detection.


\clearpage
\appendix

\thispagestyle{empty}

\onecolumn \makesupplementtitle
\section{ADDITIONAL BASELINES}
In addition to the results in the section~\ref{sec:result}, we perform more baselines with logistic regression and decision tree, which are outperformed by Random Forest. The Table~\ref{apdx:tab:hyper} and Table~\ref{apdx:tab:addition_results} show the hyperparameters and results respectively.
\begin{table*}[ht!]
\centering
\caption{Logistic Regression and Decision Tree results. }
\label{apdx:tab:addition_results}
\begin{tabular}{@{}cccc@{}}
\toprule
\multicolumn{1}{l}{}                                                                              &                                 & \begin{tabular}[c]{@{}c@{}}Logistic Regression \\ 0-1-0\end{tabular} &  \begin{tabular}[c]{@{}c@{}}Decision Tree\\ 0-1-0\end{tabular} \\ \midrule
\multicolumn{1}{c|}{\multirow{2}{*}{\begin{tabular}[c]{@{}c@{}}Transaction\\ Level\end{tabular}}} & \multicolumn{1}{c|}{Base}       &  \multicolumn{1}{c|}{0.021 ± 0.008}                    & 
\multicolumn{1}{c}{0.055 ± 0.021}                                  \\
\multicolumn{1}{c|}{}                                                                             & \multicolumn{1}{c|}{Base + Agg} & 
 \multicolumn{1}{c|}{0.029 ± 0.002}                    & 
 \multicolumn{1}{c}{0.076 ± 0.009}                                  \\ \midrule
\multicolumn{1}{c|}{\multirow{2}{*}{\begin{tabular}[c]{@{}c@{}}Card Level \\ Early\end{tabular}}}       & \multicolumn{1}{c|}{Base}       &  \multicolumn{1}{c|}{0.019 ± 0.009}                    & \multicolumn{1}{c}{0.063 ± 0.027}              \\
\multicolumn{1}{c|}{}                                                                             & \multicolumn{1}{c|}{Base + Agg} &
 \multicolumn{1}{c|}{0.029 ± 0.004}                    & \multicolumn{1}{c}{0.070 ± 0.011}              \\ 
\midrule
\multicolumn{1}{c|}{\multirow{2}{*}{\begin{tabular}[c]{@{}c@{}}Card Level \\ Regular\end{tabular}}}       & \multicolumn{1}{c|}{Base}       &  \multicolumn{1}{c|}{0.024 ± 0.010}                    & \multicolumn{1}{c}{0.075 ± 0.030}              \\
\multicolumn{1}{c|}{}                                                                             & \multicolumn{1}{c|}{Base + Agg} & 
 \multicolumn{1}{c|}{0.032 ± 0.005}                    & \multicolumn{1}{c}{0.104 ± 0.015}              \\ 

\bottomrule
\end{tabular}
\label{apdx:tab:logit_decision}
\end{table*}

\section{HYPERPARAMETERS}
\label{sec:apdx:hyper}

\begin{table*}[htb]
\centering
\caption{Hyperparameters for Grid Search, bold indicates the chosen hyperparameters.}
\label{apdx:tab:hyper}
\begin{tabular}{+ll}
\toprule 
Model & Hyperparameter \\ \otoprule
Decision Tree & max\_depth':[4,10,20,\textbf{None}]; max\_features=[1,3,\textbf{auto}];min\_samples\_leaf':[2,10,\textbf{100}];\\[2pt]
Logistic Regression & solver=[newton-cg,lbfgs, \textbf{sag}, saga]; C=[\textbf{100}, 10, 1.0, 0.1, 0.01];\\
Random Forest & bootstrap = [\textbf{True}, False]; max\_depth':[4,10,20 \textbf{None}];\\
    &min\_samples\_leaf':[2,10 \textbf{100}]; n\_estimators':[10,20,\textbf{100} ]\\
(Bi-)LSTM & lstm\_layers = [20, 150, \textbf{300}]; hidden\_layers:[20, 150, \textbf{300}]; \\
    &drop\_out:[0.2, \textbf{0.8}]; epochs:[2, 4, \textbf{8}];\\
    & learning\_rate:[$10^{-3},10^{-4},\mathbf{5.10^{-4}},5.10^{-5}$]; batch\_size:[1024, 2048, \textbf{4096}, 8192];\\[2pt]\bottomrule
\end{tabular}
\end{table*}


\begin{thebibliography}{}
\setlength{\itemindent}{-\leftmargin}
\makeatletter\renewcommand{\@biblabel}[1]{}\makeatother


\bibitem[Kou et.al, 2004]{kou}
Kou, Y., Lu, C. T., Sirwongwattana, S., and Huang, Y. P. (2004, March). Survey of fraud detection techniques. In IEEE International Conference on Networking, Sensing and Control, 2004 (Vol. 2, pp. 749-754). IEEE.
\bibitem[Pozzolo et al., 2014]{Pozzolo3}
Dal Pozzolo, Andrea, et al. "Learned lessons in credit card fraud detection from a practitioner perspective." Expert systems with applications 41.10 (2014): 4915-4928.
\bibitem[Adams et al., 2006]{adams}
Adams, G. W., Campbell, D. R., Campbell, M., \& Rose, M. P. (2006). Fraud prevention. The CPA Journal, 76(1), 56.
\bibitem[Gianini et al., 2020]{giannini}
Gianini, G., Fossi, L. G., Mio, C., Caelen, O., Brunie, L., \& Damiani, E. (2020). Managing a pool of rules for credit card fraud detection by a Game Theory based approach. Future Generation Computer Systems, 102, 549-561.
\bibitem[Le Borgne et al., 2022]{leborgne2022fraud}
Le Borgne, Y., Siblini, W., Lebichot, B. \& Bontempi, G. Reproducible Machine Learning for Credit Card Fraud Detection - Practical Handbook. (Université Libre de Bruxelles, 2022), https://github.com/Fraud-Detection-Handbook/fraud-detection-handbook
\bibitem[Awoyemi et al., 2017]{awoyemi}
Awoyemi, J. O., Adetunmbi, A. O., \& Oluwadare, S. A. (2017, October). Credit card fraud detection using machine learning techniques: A comparative analysis. In 2017 International Conference on Computing Networking and Informatics (ICCNI) (pp. 1-9). IEEE.
\bibitem[Schuster et al., 1997]{Schuster}
M. Schuster and K. K. Paliwal, "Bidirectional recurrent neural networks," in IEEE Transactions on Signal Processing, vol. 45, no. 11, pp. 2673-2681, Nov. 1997, doi: 10.1109/78.650093.
\bibitem[Siblini et al., 2020]{Siblini}
Siblini, Wissam, et al. "Master your Metrics with Calibration." International Symposium on Intelligent Data Analysis. Springer, Cham, 2020.
\bibitem[Pozzolo et al., 2015]{Pozzolo2}
Dal Pozzolo, Andrea. "Adaptive machine learning for credit card fraud detection." (2015).
\bibitem[Pozzolo et al., 2018]{Pozzolo}
Dal Pozzolo A, Boracchi G, Caelen O, Alippi C, Bontempi G. Credit Card Fraud Detection: A Realistic Modeling and a Novel Learning Strategy. IEEE Trans Neural Netw Learn Syst. 2018 Aug;29(8):3784-3797. doi: 10.1109/TNNLS.2017.2736643. Epub 2017 Sep 14. PMID: 28920909.
\bibitem[Bhattacharyya et al., 2011]{Bhattacharyya}
Bhattacharyya, Siddhartha, et al. "Data mining for credit card fraud: A comparative study." Decision support systems 50.3 (2011): 602-613.
\bibitem[Cheng et al., 2020]{Cheng}
Cheng, Dawei, et al. "Spatio-temporal attention-based neural network for credit card fraud detection." Proceedings of the AAAI Conference on Artificial Intelligence. Vol. 34. No. 01. 2020.
\bibitem[Lucas et al., 2019]{Lucas}
Lucas, Yvan, et al. "Multiple perspectives HMM-based feature engineering for credit card fraud detection." Proceedings of the 34th ACM/SIGAPP Symposium on Applied Computing. 2019.
\bibitem[Whitrow et al., 2009]{Whitrow}
Whitrow, Christopher, et al. "Transaction aggregation as a strategy for credit card fraud detection." Data mining and knowledge discovery 18.1 (2009): 30-55.
\bibitem[Bahnsen et al., 2016]{Bahnsen}
Bahnsen, Alejandro Correa, et al. "Feature engineering strategies for credit card fraud detection." Expert Systems with Applications 51 (2016): 134-142.
\bibitem[Fu et al., 2016]{Fu}
Fu, K., Cheng, D., Tu, Y., \& Zhang, L. (2016, October). Credit card fraud detection using convolutional neural networks. In International Conference on Neural Information Processing (pp. 483-490). Springer, Cham.
\bibitem[Ghosh et al., 2020]{Ghosh}
K. Ghosh Dastidar, J. Jurgovsky, W. Siblini, L. He-Guelton and M. Granitzer, "NAG: Neural Feature Aggregation Framework for Credit Card Fraud Detection," 2020 IEEE International Conference on Data Mining (ICDM), Sorrento, Italy, 2020, pp. 92-101, doi: 10.1109/ICDM50108.2020.00018.

\bibitem[Jurgovsky et al., 2018]{Jurgovsky}
Jurgovsky, Johannes, et al. "Sequence classification for credit-card fraud detection." Expert Systems with Applications 100 (2018): 234-245.
\bibitem[Ozenne et al., 2015]{Ozenne}
Ozenne B, Subtil F, Maucort-Boulch D. The precision--recall curve overcame the optimism of the receiver operating characteristic curve in rare diseases. J Clin Epidemiol. 2015 Aug;68(8):855-9. doi: 10.1016/j.jclinepi.2015.02.010. Epub 2015 Feb 28. PMID: 25881487.
\bibitem[Lebichot et al., 2020]{lebichot}
Lebichot, B., Paldino, G. M., Bontempi, G., Siblini, W., He-Guelton, L., \& Oblé, F. (2020, October). Incremental learning strategies for credit cards fraud detection. In 2020 IEEE 7th International Conference on Data Science and Advanced Analytics (DSAA) (pp. 785-786). IEEE.
\bibitem[Alazizi et al., 2019]{alazizi}
Alazizi, A., Habrard, A., Jacquenet, F., He-Guelton, L., Oblé, F., \& Siblini, W. (2019, November). Anomaly Detection, Consider Your Dataset First An Illustration on Fraud Detection. In 2019 IEEE 31st International Conference on Tools with Artificial Intelligence (ICTAI) (pp. 1351-1355). IEEE.
\bibitem[Russac et al., 2018]{Russac}
Russac, Yoan, Olivier Caelen, and Liyun He-Guelton. "Embeddings of categorical variables for sequential data in fraud context." International Conference on Advanced Machine Learning Technologies and Applications. Springer, Cham, 2018.

\bibitem[Graves et al., 2005]{Graves}
Graves, Alex, and Jürgen Schmidhuber. "Framewise phoneme classification with bidirectional LSTM and other neural network architectures." Neural networks 18.5-6 (2005): 602-610.
\bibitem[Najadat et al., 2020]{Najadat}
Najadat, Hassan, et al. "Credit card fraud detection based on machine and deep learning." 2020 11th International Conference on Information and Communication Systems (ICICS). IEEE, 2020.
\bibitem[Cui et al., 2018]{Cui}
 Cui, Zhiyong, et al. "Deep bidirectional and unidirectional LSTM recurrent neural network for network-wide traffic speed prediction." arXiv preprint arXiv:1801.02143 (2018).
 
 \bibitem[Bengio et al., 1994]{Bengio}
 Y. Bengio, P. Simard and P. Frasconi, "Learning long-term dependencies with gradient descent is difficult," in IEEE Transactions on Neural Networks, vol. 5, no. 2, pp. 157-166, March 1994, doi: 10.1109/72.279181.

\bibitem[Nowak et al., 2017]{Nowak}
Nowak, J., Taspinar, A., and Scherer, R. (2017, June). LSTM recurrent neural networks for short text and sentiment classification. In International Conference on Artificial Intelligence and Soft Computing (pp. 553-562). Springer, Cham.

\bibitem[Althelaya et al., 2018]{Althelaya}
 Althelaya, Khaled A., El-Sayed M. El-Alfy, and Salahadin Mohammed. "Evaluation of bidirectional LSTM for short-and long-term stock market prediction." 2018 9th international conference on information and communication systems (ICICS). IEEE, 2018.
 
 \bibitem[Hochreiter et al., 1997]{Hochreiter2}

 Hochreiter, Sepp, and Jürgen Schmidhuber. "Long short-term memory." Neural computation 9.8 (1997): 1735-1780.

\bibitem[Persio et al., 2017]{Persio}
Di Persio, Luca, and Oleksandr Honchar. "Analysis of recurrent neural networks for short-term energy load forecasting." AIP Conference Proceedings. Vol. 1906. No. 1. AIP Publishing LLC, 2017.

\bibitem[Kingma et al., 2014]{kingma2014adam}
Kingma, Diederik P., and Jimmy Ba. "Adam: A method for stochastic optimization." arXiv preprint arXiv:1412.6980 Add to Citavi project by ArXiv ID (2014).
\bibitem[Saito et al., 2015]{SR15}
Saito, Takaya, and Marc Rehmsmeier. "The precision-recall plot is more informative than the ROC plot when evaluating binary classifiers on imbalanced datasets." PloS one 10.3 (2015): e0118432.

\bibitem[Li et al., 2019]{li2019}
Li, Longfei, et al. "A Time Attention based Fraud Transaction Detection Framework." arXiv preprint arXiv:1912.11760 (2019).
\end{thebibliography}
\end{document}